\documentclass{article}

\usepackage[nonatbib,final]{nips_2017}

\usepackage[utf8]{inputenc} 
\usepackage[T1]{fontenc}    
\usepackage{hyperref}       
\usepackage{url}            
\usepackage{booktabs}       
\usepackage{amsfonts}       
\usepackage{nicefrac}       
\usepackage{microtype}      
\usepackage{placeins}

\usepackage{graphicx}
\usepackage[margin=1cm]{caption}
\usepackage{color}
\usepackage{tikz}
\usetikzlibrary{arrows, automata, backgrounds, calendar, chains,
decorations, hobby,intersections, matrix, mindmap, patterns, petri,
positioning, shadows, shapes, shapes.geometric,trees,shadings,
decorations.markings,snakes}

\newcommand{\X}{\mathcal{X}}
\newcommand{\Y}{\mathcal{Y}}
\newcommand{\Z}{\mathcal{Z}}
\newcommand{\R}{\mathbb{R}}
\newcommand{\D}{\mathcal{D}}
\newcommand{\C}{\mathcal{C}}
\newcommand{\x}{\vec{x}}

\renewcommand{\vec}[1]{\mathbf{#1}}

\newtheorem{remark}{Remark}

\title{Forward Thinking: Building and Training Neural Networks One Layer at a Time}

\author{
Chris Hettinger, Tanner Christensen, Ben Ehlert, \\
\textbf{Jeffrey Humpherys, Tyler Jarvis, and Sean Wade}\\
Department of Mathematics\\ 
Brigham Young University\\
Provo, Utah 84602 \\
\texttt{hettinger@math.byu.edu, tannerkchristensen@gmail.com,}\\
\texttt{benjaminwehlert@gmail.com, jeffh@math.byu.edu,} \\
\texttt{jarvis@math.byu.edu,seanwademail@gmail.com}
}

\begin{document}

\maketitle

\begin{abstract}
We present a general framework for training deep neural networks without backpropagation.  
This substantially decreases training time and also allows for construction of deep networks with many sorts of learners, including networks whose layers are defined by functions that are not easily differentiated, like decision trees. 
The main idea is that layers can be trained one at a time, and once they are trained, the input data are mapped forward through the layer to create a new learning problem.  The process is repeated, transforming the data through multiple layers, one at a time, rendering a new data set, which is expected to be better behaved, and on which a final output layer can achieve good performance.  We call this \emph{forward thinking} and demonstrate a proof of concept by achieving state-of-the-art accuracy on the MNIST dataset for convolutional neural networks.  We also provide a general mathematical formulation of forward thinking that allows for other types of deep learning problems to be considered.
\end{abstract}

\section{Introduction}

In recent years, deep neural networks (DNNs) have become a dominant force in many supervised learning problems.  In several side-by-side comparisons with standardized data sets and well-defined benchmarks, neural networks  have bested most and in some cases all other leading machine learning techniques \cite{Krizhevsky,socher:2013}.  This is particularly pronounced in image, speech, and natural language recognition problems, where deep learning methods are also consistently beating humans \cite{betterthanhumans}.

The main downside of deep learning is the computational complexity of the training algorithms.  In particular, it is extremely expensive computationally to use backpropagation to train multiple layers of nonlinear activation functions \cite{init_and_momentum}.  Indeed, the computational resources required to fully train a DNN are in many cases orders of magnitude greater  than other machine learning methods that perform almost as well on many tasks \cite{comparo}.  In other words, in many cases a great deal of extra work is required to get only slightly better performance.  

We present a general framework, 
which we call \emph{forward thinking}, 
for training DNNs without doing backpropagation. This allows a network to be built from scratch as 
deep as needed in real time.  
It also allows the use of many different sorts of learners in the layers of the network, including learners that are not easily differentiable, like random forests.  
The main idea is that layers of learning functions can be trained one at a time, and once trained, the input data can be mapped forward through the layer to create a new learning problem.  This  process is repeated, transforming the data through multiple layers, and rendering a new data set, which is expected to be better behaved, and on which a final output layer can achieve good performance.  
This is much faster than traditional backpropagation, and the number of layers can be determined at training time by simply continuing to add and train layers consecutively until performance plateaus.

This greedy approach to deep learning stems from a confluence of ideas that generalize nicely into a single framework.  In particular several recent papers have elements that can be nicely described as special cases of forward thinking; see for example, net2net \cite{net2net}, cascade correlation \cite{cascade}, network morphism \cite{morphism1,morphism2}, and convolutional representation transfer \cite{6909618}.  
Many variants of the ideas behind forward thinking have been proposed in various settings \cite{Rueda-Plata2015, Roy, Goodfellow-Bengio-book, Bengio-greedy, Larochelle, Wu}.   But the full potential of this idea does not seem to have been completely realized.  Specifically, the papers \cite{Bengio-greedy} and \cite{Larochelle} did some experiments training networks in a greedy fashion and saw poor performance.  Others who have used greedy training methods have used them only for pretraining, that is, as a method for initializing networks that are subsequently trained using backpropagation.

However, we show here that forward thinking can be effective as a stand-alone training method.  It is, of course, much faster than backpropagation, and yet it can give results that are as accurate as backpropagation. 
As a proof of concept, we use forward thinking to design and train both a fully-connected deep neural network (DNN) and a convolutional neural network (CNN) and compare their performance on the MNIST dataset against their traditionally trained counterparts.  In an ``apples-to-apples'' comparison against traditionally trained networks, we find roughly equivalent performance in terms of accuracy and significantly reduced training time.  In particular, we were able to get an accuracy of 98.89\% with a forward-thinking DNN and a state-of-the-art accuracy of 99.72\% with a forward-thinking CNN.

The rest of this paper describes how forward thinking can be used to build (both fully connected and convolutional) feedforward neural networks one layer at a time.  In a companion paper, we consider deep random forests and show that many of the ideas presented here carry over to other machine learning techniques \cite{ft_forests}.  Specifically, in that paper we replace neurons with decision trees and describe a specific implementation, which, as a proof of concept, also achieves very good results on the MNIST dataset.
Together these two papers illustrate how the forward thinking framework can be applied to many machine learning methods.

\section{General mathematical description of forward thinking} \label{architecture}
\label{sec:math-description}

In this section, we describe the general mathematical structure of a forward thinking deep network.  The main idea is that neurons can be generalized to any type of learner and then, once trained, the input data are mapped forward through the layer to create a new learning problem.  This process is repeated, transforming the data through multiple layers, one at a time, and rendering a new, which is expected to be better behaved, and on which a final output layer can achieve good performance.

\subsection*{The input layer}

The data $\D^{(0)} = \{(\x_i^{(0)},y_i)\}^N_{i=1} \subset \X^{(0)}\times \Y$ are given as the set of input values $\vec{x}_i^{(0)}$ from a set $\X^{(0)}$ and their corresponding outputs $y_i$ in a set $\Y$.

In many learning problems, $\X^{(0)} \subset \mathbb{R}^d$, which means that there are $d$ real-valued features. If the inputs are images, we can stack them as large vectors where each pixel is a component.  In some deep learning problems, each input is a stack of images, for example color images can be represented as three separate monochromatic images, or three separate channels of the image.

For binary classification problems, the output space can be taken to be $\Y= \{-1,1\}$.  For multi-class problems we often set $\Y= \{1,2,\ldots,n\}$.

\subsection*{The first hidden layer}

Let $\C^{(1)} = \{ C_1^{(1)}, C_2^{(1)},\ldots, C_{m_1}^{(1)}\}$ be a  layer  of $m_1$ learning functions $C_i^{(1)}:\X^{(0)}\to \Z_i^{(1)}$, for some codomain $\Z_i^{(1)}$, with parameters $\theta_i^{(1)}$.   These learning functions (or learners) can be regression, classification, or kernel functions, and can be thought of as defining new features, as follows. Let $\X^{(1)} = \Z_1^{(1)}\times \Z_2^{(1)}\times\cdots\times \Z_{m_1}^{(1)}$ and transfer the inputs $\{\vec{x}_i^{(0)}\}^N_{i=1}\subset\X^{(0)}$ to $\X^{(1)}$ according to the map
\[
\vec{x}_i^{(1)} = (C_1^{(1)}(\vec{x}^{(0)}_i), C_2^{(1)}(\vec{x}^{(0)}_i),\ldots, C_m^{(1)}(\vec{x}^{(0)}_i)) \subset \X^{(1)},\quad i=1,\ldots,N.
\]
This gives a new data set $\mathcal{D}^{(1)} = \{(\vec{x}_i^{(1)},y_i)\}^N_{i=1}\subset \X^{(1)} \times \Y$.

In many learning problems $\Z^{(1)} = [-1,1]$, in which case the new domain $\X^{(1)}$ is a hypercube $[-1,1]^{m_1}$.  It is also common to have $\Z^{(1)}=[0,\infty)$, in which case $\X^{(1)}$ is the $m_1$-dimensional orthant $[0,\infty)^{m_1}$.

The goal is to choose $\C^{(1)}$ to make the new dataset ``more separable,'' or better behaved, in some sense, than the previous dataset.  As we repeat this process iteratively, the data should become increasingly well behaved, so that in the final layer, a single learner can finish the job.

The functions $\C^{(1)}$ are trained on the data set $\D^{(0)} = \{(\x_i^{(0)},y_i)\}^N_{i=1}$ in some suitable way.  In many settings, this could be done by minimizing a loss function for a neural network with a single hidden layer $\C^{(1)}$ and a final output layer consisting of just one learner $C':\X^{(1)} \to \R$ with parameters $\theta'$. The loss function could then be of the form
\[
L^{(1)}(\Theta^{(1)},\theta') = \sum_{i=1}^N \ell(C'\circ \C^{(1)}(\x_i), y_i) + r(\Theta^{(1)}, \theta'),
\]
where $\ell:\R\times \Y \to [0,\infty)$ is a measure of how close $C'\circ \C^{(1)}(\x_i)$ is to $y_i$ and $r(\Theta^{(1)}, \theta')$ is a regularization term.  Of course, the loss function for training this layer need not be of this form, but this would be an obvious choice.  If the learners $\C^{(0)}$ are regression trees or random forests, these could be trained in the standard way, without the extra learner $C'$.

The key point is that once they are learned, the parameters $\theta_i^{(1)}$ are frozen, but the parameters $\theta'$ are discarded (and the learner $C'$ may either be discarded or retrained at the next iteration).  The old data $\x_i^{(0)}$ are mapped through the learned layer $\C^{(1)}$ to give new data $\x_i^{(1)}$, which is then passed to a new hidden layer.

\subsection*{Additional hidden layers}

Let $\C^{(k)} = \{ C_1^{(k)}, C_2^{(k)},\ldots, C_{m_k}^{(k)}\}$ be a set (layer) of $m_k$ learning functions $C_i^{(k)}:\X^{(k-1)}\to \Z^{(k)}$.  This layer is again trained on the data $\D^{(k-1)} = \{(\x_i^{(k-1)},y_i)\}$. This would usually be done in the same manner as the previous layer, but it need not be the same; for example, if the new layer consists of a different kind of learners, then the training method for the new layer might also need to differ.

As with the first layer, the inputs $\{\x_i^{(k-1)}\}^N_{i=1}\subset \X^{(k-1)} = \Z_1^{(k-1)}\times \Z_2^{(k-1)}\times\cdots\times \Z_{m_{k-1}}^{(k-1)}$ are transferred to a new domain $\{\x_i^{(k)}\}^N_{i=1} \subset \X^{(k)} = \Z_1^{(k)}\times \Z_2^{(k)}\times\cdots\times \Z_{m_k}^{(k)}$ according to the map
\[
\x_i^{(k)} = (C_1^{(k)}(\x_i^{(k-1)}), C_2^{(k)}(\x_i^{(k-1)}),\ldots, C_{m_k}^{(k)}(\x_i^{(k-1)})),\quad i=1,\ldots,N.
\]
This gives a new data set $\mathcal{D}^{(k)} = \{(\x_i^{(k)},y_i)\}^N_{i=1}\subset \X^{(k)} \times \Y$, and the process is repeated.

\subsubsection*{Final layer}

After passing the data through the last hidden layer $\mathcal{D}^{(n)} = \{(\x_i^{(n)},y_i)\}^N_{i=1}\subset \X^{(n)}\times\Y$ we train a final layer, which consists of a single learning function $C_F:\X^{(n)}\to \Y$, to determine the outputs, where the $C_F(\x_i^{(n)})$ is expected to be close to $y_i$ for each $i$.

\section{Building fully-connected networks}
We now explain how to implement Forward Thinking to build a fully-connected DNN.  We call the result a \emph{Forward Thinking Deep Neural Network} or \emph{FTDNN}.

\begin{figure}[h]
\centering
\resizebox{0.8\linewidth}{!}{
\begin{tikzpicture}[doubRect/.style={rectangle,very thick,double,double distance=1mm,text width=6em,text height=2ex, text depth=.25ex, align=center,minimum height=20pt},
rect/.style={rectangle,very thick,text width=6em,text height=2ex, text depth=.25ex, align=center,minimum height=20pt},
doubEll/.style={ellipse,very thick,double,double distance=1mm,text width=4em,text height=2ex,text depth=.25ex,align=center},
triangle/.style={black!35!blue,very thick,regular polygon,regular polygon sides=3}]
\node[doubEll,draw,very thick,font=\large] (x1) {$\mathcal{D}^{(0)}$};
\node[rect,draw,very thick,font=\large,black!35!blue,below=.5cm of x1] (D11) {$\mathcal{C}^{(1)}$};
\node[triangle,draw,very thick,font=\large,shape border rotate=180,below=.5cm of D11] (F1) {$C_F$};
\node[doubEll,draw,very thick,font=\large,right=3cm of x1] (x2) {$\mathcal{D}^{(0)}$};
\node[rect,draw,very thick,font=\large,doubRect,below=.5cm of x2] (D12) {$\mathcal{D}^{(1)}$};
\node[rect,draw,very thick,font=\large,black!35!blue,below=.5cm of D12] (D21) {$\mathcal{C}^{(2)}$};
\node[triangle,draw,very thick,font=\large,shape border rotate=180,below=.5cm of D21] (F2) {$C_F$};
\node[doubEll,draw,very thick,font=\large,right=3cm of x2] (x3) {$\mathcal{D}^{(0)}$};
\node[rect,draw,very thick,font=\large,doubRect,below=.5cm of x3] (D13) {$\mathcal{D}^{(1)}$};
\node[rect,draw,very thick,font=\large,doubRect,below=.5cm of D13] (D22) {$\mathcal{D}^{(2)}$};
\node[rect,draw,very thick,font=\large,black!35!blue,below=.5cm of D22] (D3) {$\mathcal{C}^{(3)}$};
\node[triangle,draw,very thick,font=\large,shape border rotate=180,below=.5cm of D3] (F3) {$C_F$};
\foreach \s/\t in {D11/F1,D21/F2,D3/F3}
{\draw[very thick,black!35!blue] (\s)--(\t);}
\foreach \s/\t in {x1/D11,D12/D21,D22/D3}
{\draw[very thick,black!35!blue,shorten <=.06cm] (\s)--(\t);}
\node[draw=none,below right=.3cm and .3cm of D11] (Aux1) {};
\node[draw=none,below left=.3cm and .3cm of D12] (Aux2) {};
\node[draw=none,below right=.3cm and .3cm of D12] (Aux3) {};
\node[draw=none,below left=.3cm and .3cm of D13] (Aux4) {};
\draw[->,>=stealth',very thick,green!60!black] (Aux1) -- (Aux2);
\draw[->,>=stealth',very thick,green!60!black] (Aux3) -- (Aux4);
\foreach \s/\t in {x2/D12,x3/D13,D13/D22}
{\draw[double,double distance=.1cm,very thick,shorten <=.06cm,shorten >=.06cm] (\s)--(\t);}
\end{tikzpicture}}
\caption{The first three iterations of a fully-connected network built with the forward thinking algorithm. The original data set is represented by an ellipse, fully-connected layers with rectangles, and final (output) layers with triangles. Layers with single blue outlines are trainable, while those with double black outlines have been frozen and thus turned into new data sets.}
\end{figure}
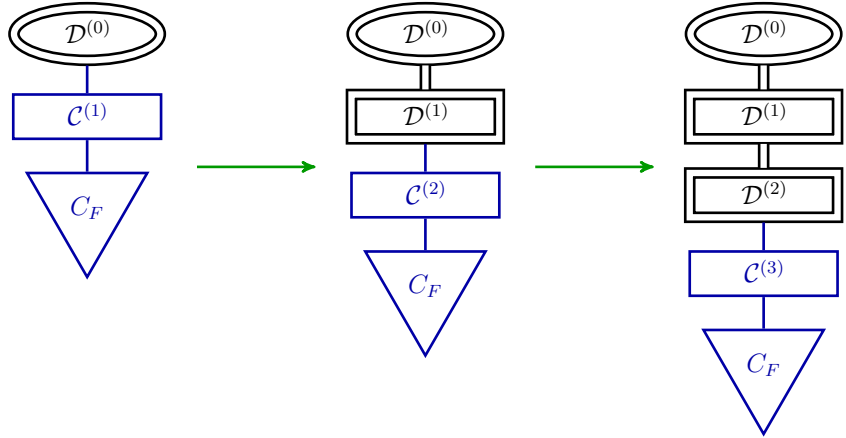

\subsection{The base network}
As with any neural network, we begin by selecting an output layer and loss function appropriate to the problem at hand. For example, a binary classification problem might call for an output layer consisting of a single neuron with a sigmoid activation function and a categorical cross-entropy loss function.

Then we construct a network with a single hidden layer of appropriate width, using conventional activation functions, such as ReLU.  We also randomly initialize the parameters of this network. Tools like weight regularization and dropout can be used during training.

In our setting, it does not pay to train this first single-hidden-layer network too long. Instead of milking this first layer for incremental  improvements, one can make bigger improvements by moving on to the next step, adding another layer.

\subsection{Freezing the hidden layer}

Once the first network is trained, the weights coming into the first layer are frozen (and stored), and the training inputs $\{\x^{(0)}_i\}_{i=1}^N$ are pushed through the resulting layer to give new ``synthetic'' data $\{\x^{(1)}_i\}_{i=1}^N$, which is used to train the next layer. The weights for the output layer are discarded (they will be retrained at each step).  

The main advantages of freezing the previously trained weights are ($i$) speed: adding each new layer amounts to training a shallow network with only one hidden layer; and ($ii$) resilience to overfitting.

\subsection{Adding a new layer}
Now insert a new hidden layer between the previously trained layer and the output layer. This layer is trained as a single-hidden-layer network on the new, synthetic data $\{\x_i^{(1)}\}_{i=1}^N$ constructed at the previous step.  Randomly initialize the parameters of this layer and randomly re-initialize those of the output layer. This will cause a temporary dip in the performance of the network, but it also creates new room for improvement.

\subsection{Iterating}

The process of freezing old layers and inserting new ones is repeated until additional layers cease to improve performance. This indicates that it's time to stop adding new layers and consider the network complete.

Even though each stage involves training only a shallow network, the layers together form a single deep network.  As mentioned before, the main advantage of this method is that we never need to use backpropagation to reach back into the network and train deep parameters. So we avoid the pitfalls of backpropagation, including its high computational cost and its struggle to effectively adjust deep parameters.

\begin{remark} Throughout this paper we use the term \emph{backpropagation} informally to mean the traditional process of training a DNN by some variant of stochastic gradient descent (SGD), combined with the chain rule and cascading derivatives.  In this paper, when we train each of our shallow intermediate networks, we do still use SGD at each stage, but we do not consider this an instance of backpropagation, because there are no long chains of cascading derivatives to calculate.  
\end{remark}

\subsection{Fully-connected network results}

We used both forward thinking and traditional backpropagation to train a fully-connected network with four hidden layers of 150, 100, 50, and 10 neurons respectively, applied to the MNIST handwritten digit dataset (we also  followed the common practice of augmenting the training set by slightly scaling, rotating, and shifting the images).  

Test accuracy was comparable between the the networks trained using these two methods, but training with forward thinking was significantly faster, as described below.  

As explained earlier, training using forward thinking means that we started with a network of only one hidden layer of 150 neurons. After that layer was trained, the data was pushed through the trained layer to produce a new 150-dimensional synthetic data set. Then we trained a new hidden layer with 100 neurons on the new data set on a layer, and then repeated the process for a hidden layer of 50. This network used a 10-neuron softmax output layer and a categorical-cross entropy loss function.

To have a benchmark for forward thinking, we trained a DNN of identical architecture in the conventional way, by optimizing all of the parameters at once with backpropagation. We tuned hyperparameters such as learning rate and regularization constants separately for the forward-thinking-trained DNN and the traditionally trained DNN so as to maximize the performance of each and provide a fair comparison.

\begin{figure}[h]
\begin{center}$
\begin{array}{cc}
\includegraphics[height=2.7in]{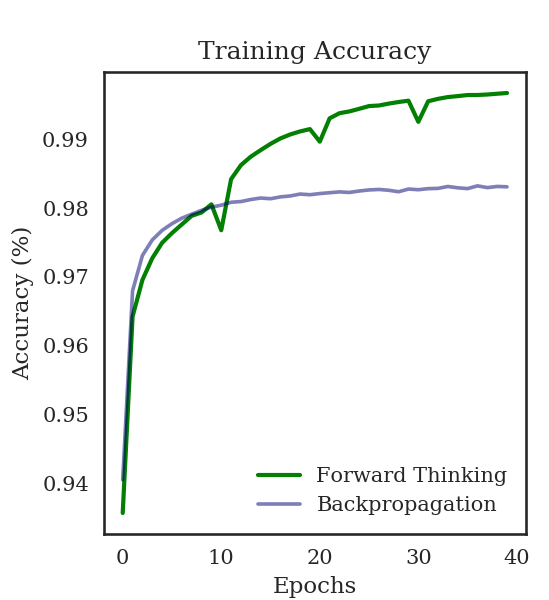} & \includegraphics[height=2.7in]{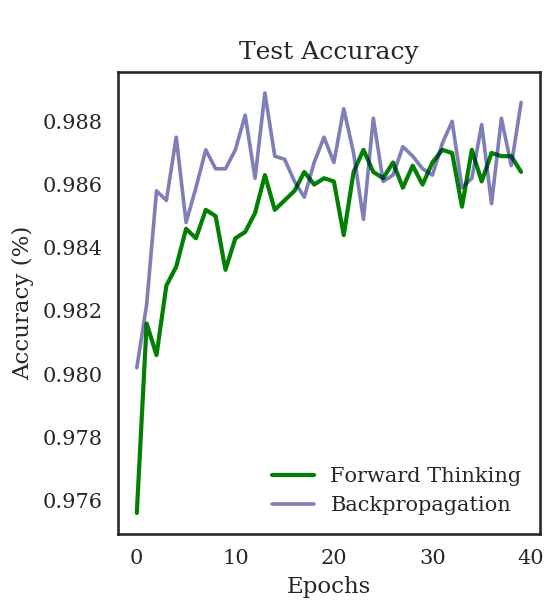} \\ 
$\quad\qquad(a)$ & $\quad\qquad(b)$
\end{array}$
\end{center}
\caption{A comparison of the training and test accuracy per epoch of a convolutional neural network trained using forward thinking (thicker, green) and traditional backpropagation (thinner, blue).  Notice that $(a)$ forward thinking fit the network quickly and precisely to the training data, while training with backpropagation leveled off at lower accuracy.  The brief dips in training accuracy for forward thinking occur when a new layer is added.  Also, $(b)$ the final testing accuracy was similar for both methods, with backprop retaining a slight edge, but the time to train each epoch, and the overall training time were both much faster for forward thinking.}
\label{fig:aug-mnist}
\end{figure}

The forward thinking network fit itself quickly and precisely to the training data, while training with backpropagation leveled off, as shown in Figure~\ref  {fig:aug-mnist}$(a)$. This suggests that more can be done to prevent overfitting in forward thinking networks. The forward thinking network suffered dips in training performance when adding new layers, but quickly recovered. 

On the same machine, overall training time for forward thinking was about 30\% faster than backpropagation. This speedup occurred in spite of the fact that both were trained using libraries optimized for backpropagation. We expect that custom code would increase this speed advantage. 

Testing accuracy was similar for both methods, with backpropagation retaining a slight edge, as shown in Figure~\ref{fig:aug-mnist}$(b)$. This reinforces the idea that anti-overfitting methods for forward thinking nets could be improved.  

\section{Convolutional networks}
We can also build convolutional networks with forward thinking. In this case we start with two hidden layers---one convolutional and one fully-connected---
 At each subsequent iteration we add a new convolutional layer before the fully connected layer at the end.  We freeze the previous convolutional layer at each step but do not freeze the fully connected layer. Convolutional tools such as max pooling can also be used in this process.

\begin{figure}[h]
\centering
\resizebox{0.7\linewidth}{!}{
\begin{tikzpicture}[scale=1,doubRect/.style={rectangle,very thick,double,double distance=1mm,text width=6em,text height=2ex, text depth=.25ex, align=center,minimum height=20pt},
rect/.style={rectangle,very thick,text width=6em,text height=2ex, text depth=.25ex, align=center,minimum height=20pt},
doubEll/.style={ellipse,very thick,double,double distance=1mm,text width=4em,text height=2ex,text depth=.25ex,align=center},
triangle/.style={black!35!blue,very thick,regular polygon,regular polygon sides=3},
doubDiam/.style={diamond,very thick,double,double distance=1mm,text width=2em,text height=2ex, text depth=.25ex, align=center,minimum height=20pt},]
\node[doubEll,draw,font=\large] (x1) {$\mathcal{D}^{(0)}$};
\node[diamond,draw,font=\large,black!35!blue,very thick,below=.5cm of x1] (C11) {$\mathcal{C}^{(1)}$};
\node[rect,draw,font=\large,black!35!blue,very thick,below=.5cm of C11] (D1) {$\mathcal{C}^{(2)}$};
\node[triangle,draw,font=\large,shape border rotate=180,below=.5cm of D1] (F1) {$C_F$};
\node[doubEll,draw,font=\large,right=3cm of x1] (x2) {$\mathcal{D}^{(0)}$};
\node[diamond,draw,font=\large,doubDiam,very thick,below=.5cm of x2] (C12) {$\mathcal{D}^{(1)}$};
\node[diamond,draw,font=\large,black!35!blue,very thick,below=.5cm of C12] (C21) {$\mathcal{C}^{(2)}$};
\node[rect,draw,font=\large,black!35!blue,very thick,below=.5cm of C21] (D21) {$\mathcal{C}^{(3)}$};
\node[triangle,draw,font=\large,shape border rotate=180,below=.5cm of D21] (F2) {$C_F$};
\node[doubEll,draw,font=\large,right=3cm of x2] (x3) {$\mathcal{D}^{(0)}$};
\node[diamond,draw,font=\large,doubDiam,very thick,below=.5cm of x3] (C13) {$\mathcal{D}^{(1)}$};
\node[diamond,draw,font=\large,doubDiam,very thick,below=.5cm of C13] (C22) {$\mathcal{D}^{(2)}$};
\node[diamond,draw,font=\large,black!35!blue,very thick,below=.5cm of C22] (C3) {$\mathcal{C}^{(3)}$};
\node[rect,draw,font=\large,black!35!blue,very thick,below=.5cm of C3] (D22) {$\mathcal{C}^{(4)}$};
\node[triangle,draw,font=\large,shape border rotate=180,below=.5cm of D22] (F3) {$C_F$};
\foreach \s/\t in {C11/D1,D1/F1,C21/D21,D21/F2,C3/D22,D22/F3}
{\draw[very thick,black!35!blue] (\s)--(\t);}
\foreach \s/\t in {x1/C11,C12/C21,C22/C3}
{\draw[very thick,black!35!blue,shorten <=.06cm] (\s)--(\t);}
\node[draw=none,below right=.3cm and .3cm of C11] (Aux1) {};
\node[draw=none,below left=.3cm and .3cm of C12] (Aux2) {};
\node[draw=none,below right=.3cm and .3cm of C12] (Aux3) {};
\node[draw=none,below left=.3cm and .3cm of C13] (Aux4) {};
\draw[->,>=stealth',very thick,green!60!black] (Aux1) -- (Aux2);
\draw[->,>=stealth',very thick,green!60!black] (Aux3) -- (Aux4);
\foreach \s/\t in {x2/C12,x3/C13,C13/C22}
{\draw[double,double distance=.1cm,very thick,shorten <=.06cm,shorten >=.06cm] (\s)--(\t);}
\end{tikzpicture}}
\caption{The first three iterations of a convolutional network built with the forward thinking algorithm. The original data set is represented by an ellipse, convolutional layers with diamonds, fully-connected layers with rectangles, and final (output) layers with triangles. Layers with single blue outlines are trainable, while those with double black outlines have been frozen and thus turned into new data sets.}
\end{figure}
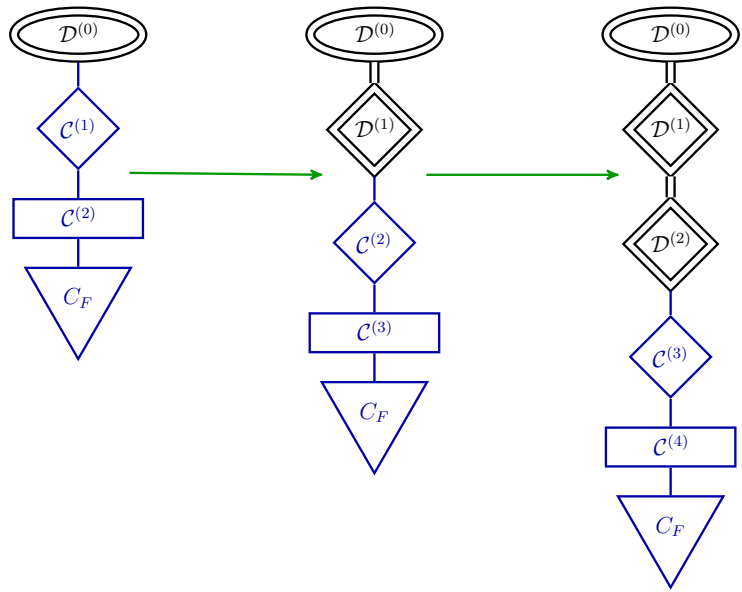

\subsection{Convolutional network results}
We used both forward thinking and backpropagation to train a convolutional neural network on the augmented MNIST dataset (augmented, as before, with slightly rotated, shifted, and scaled versions of the original training images).  

The underlying architecture of the network consists of two identical layers of 256 $3\times 3$ convolutions, with maxpooling, followed by a layer of 128 $3\times3$ convolutions, and then a fully connected layer of 150 neurons, and then final 10-class softmax layer (Softmax 10).  We trained each network (forward thinking and backpropagation) for 100 epochs (complete passes through the data.)

To train this using forward thinking we first train 256 3x3 convolutions along with a fully-connected layer of 150 ReLU neurons (FC 150) and a final 10-class softmax layer (Softmax 10). For the second iteration, we begin by pushing the data through the 256 convolutions to create a new synthetic dataset. Using this transformed data, we train an identical network: 256 3x3 convolutions followed by FC 150 and Softmax 10. As before, we push the data through the newly learned filters. For both of these first iterations, we train with an aggressive learning rate for only one epoch. With our new dataset (the original data having been passed through both sets of 256 convolutions), we learn a similar network architecture: 128 3x3 Conv, FC 150, Softmax 10. In each of the 3 iterations, we use a 2x2 max pool and a dropout of $0.3$ immediately before FC 150. Additionally, we include a dropout of $0.5$ in between FC 150 and Softmax 10.

\begin{figure}[h]
\begin{center}$
\begin{array}{cc}
\includegraphics[height=2.7in]{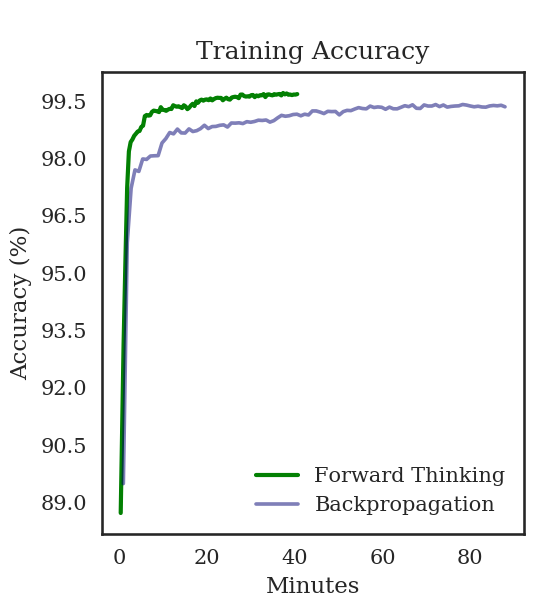} & \includegraphics[height=2.7in]{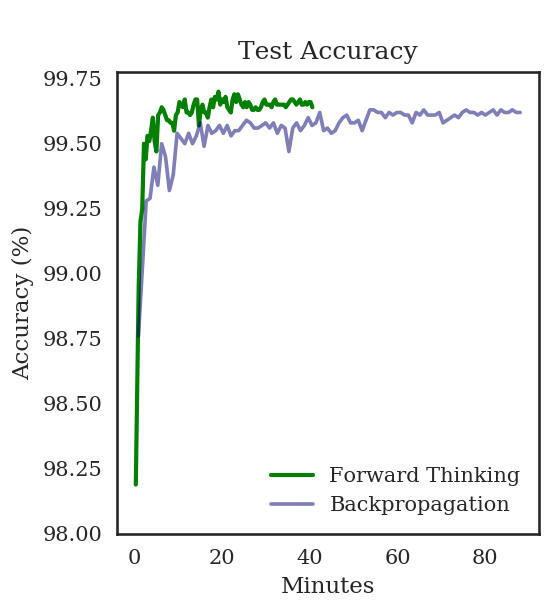} \\ 
$\quad\qquad(a)$ & $\quad\qquad(b)$
\end{array}$
\end{center}
\caption{A comparison of the training and test accuracy per minute of training time of a convolutional neural network trained using forward thinking (thicker, green) and traditional backpropagation (thinner, blue).  We have plotted accuracy against time rather than against epoch, so the two curves do not span the same horizontal length.  Notice that both $(a)$ the train and $(b)$ test accuracy for forward thinking quickly attain a level that backpropagation never reaches. }
\label{fig:conv}
\end{figure}

As shown in Figure \ref{fig:conv}, the forward-thinking-trained network out-performed the identical CNN architecture trained using backpropagation. 
Notice that both the train and test accuracy for the forward thinking net quickly attains a level that the backpropagation net never reaches.
In fact, our forward thinking CNN achieves near state-of-the-art performance (single classifier) of 99.72\% accuracy. At the time of this writing, this was the 5th ranked result according to \cite{topmnist}.

Our experiments were run on a single desktop with an Intel i5-7400 processor and an Nvidia GeForce GTX 1060 3GB GPU. We ran both our forward thinking neural network and the backprop neural network for 100 epochs. Our forward thinking neural network trained with a rate of 24 sec per epoch. Traditional backprop took 53 sec per epoch. However, this doesn't properly illustrate how much faster forward thinking should be. Because the computations above were done using libraries optimized for backpropagation, there was still a lot of unnecessary overhead. Improvements to the implementation should make forward thinking many times faster than backpropagation, and the advantage should grow with the depth of the network.

\section{Related work}

\subsection{Knowledge transfer and network morphisms}

Goodfellow et al.~outline two methods of transferring knowledge from one network to another that is deeper, wider, or both \cite{net2net}. Wei et al.~transform one network into another with a different architecture and presents mathematical formulas for doing so \cite{morphism1}\cite{morphism2}. As in the work of Goodfellow et al., this allows a new network to pick up where a previous leveled off and improved from there. Oquab et al.~take a neural network trained to classify one set of images and transfer the mid-level representations learned by its convolutional layers to a new network to be used on new images \cite{6909618}. Even when two image sets are quite different, starting the second network off with these representations increases its performance.

Although similar to these knowledge-transfer methods in some ways, forward thinking differs in that instead of training one network with backpropagation, transferring its knowledge to a deeper network, and then retraining the entire new network with backpropagation, forward thinking builds a network one layer at a time, from scratch, and does not retrain previously trained layers. At each stage, we could frame the process of adding a new layer as transferring knowledge from one network to a deeper one, but freezing old layers gives  significant benefits in speed and resistance to overfitting.

\subsection{Cascade correlation}

Cascade correlation was an early neural network algorithm that effectively let neural networks design themselves by adding and training a single neuron at a time \cite{cascade}. New neurons were added alongside the features of the data set, much as kernels are added on as new features in other machine leaning models such as support vector machines. Cascade correlation can be considered as an early indication of the potential of forward thinking.

Rather than train a single neuron at a time, we train layers, and rather than feeding old data to new layers, we only train new layers on the new, synthetic data from the previous layer.

\subsection{Greedy Pretraining}

Many others have proposed or used various greedy methods for pretraining deep neural networks \cite{Rueda-Plata2015, Roy, Goodfellow-Bengio-book, Bengio-greedy, Larochelle, Wu}.  We note that \cite{Bengio-greedy} and \cite{Larochelle} did some experiments training networks in a greedy fashion similar to forward thinking and saw poor performance.  Presumably because of those poor preliminary results, others who have used greedy training methods have used them only for initializing networks that are subsequently trained using backpropagation.  


\subsection*{Reproducibility}
All of the code used to produce our results is available in our github repository at https://github.com/tkchris93/ForwardThinking.

\subsubsection*{Acknowledgments}
This work was supported in part by the National Science Foundation, Grants number 1323785 and 1564502 and by the Defense Threat Reduction Agency, Grant Number HDRTA1-15-0049.

\bibliographystyle{plain}
\bibliography{references}

\end{document}